# Machine Learning-Based Tea Leaf Disease Detection: A Comprehensive Review

Faruk Ahmed
Research Student
Department of Computer Science and Engineering
Daffodil International University, Dhaka, Bangladesh
faruk15-4205@diu.edu.bd

Dr. Md. Taimur Ahad
Associate Professor
Department of Computer Science and Engineering
Daffodil International University, Dhaka, Bangladesh
taimurahad.cse@diu.edu.bd

Yousuf Rayhan Emon
Teaching Assistant
Department of Computer Science and Engineering
Daffodil International University, Dhaka, Bangladesh
Yousuf15-3220@diu.edu.bd

***Abstract:*** *Tea leaf diseases are a major challenge to agricultural productivity, with far-reaching implications for yield and quality in the tea industry. The rise of machine learning has enabled the development of innovative approaches to combat these diseases. Early detection and diagnosis are crucial for effective crop management. For predicting tea leaf disease, several automated systems have already been developed using different image processing techniques. This paper delivers a systematic review of the literature on machine learning methodologies applied to diagnose tea leaf disease via image classification. It thoroughly evaluates the strengths and constraints of various Vision Transformer models, including Inception Convolutional Vision Transformer (ICVT), GreenViT, PlantXViT, PlantViT, MSCVT, Transfer Learning Model & Vision Transformer (TLMViT), IterationViT, IEM-ViT. Moreover, this paper also reviews models like Dense Convolutional Network (DenseNet), Residual Neural Network (ResNet)-50V2, YOLOv5, YOLOv7, Convolutional Neural Network (CNN), Deep CNN, Non-dominated Sorting Genetic Algorithm (NSGA-II), MobileNetv2, and Lesion-Aware Visual Transformer. These machine-learning models have been tested on various datasets, demonstrating their real-world applicability. This review study* *not only highlights current progress in the field but also provides valuable insights for future research directions in the machine learning-based detection and classification of tea leaf diseases.*

## Introduction:

Vision Transformer (VIT) is a deep-learning neural network architecture for image classification (Alzahrani et al., 2023). Vision Transformer (ViT) is a detection method based on pattern recognition and deep learning that can automatically fit image features and use features to classify and predict images (Fu et al., 2023). Transformer architecture development is at the forefront of natural language processing (NLP) innovation. The ViT architecture for image detection was inspired by the effectiveness of self-attention-based deep neural networks in transformer models used for natural language processing (Alshammari et al., 2022). However, Convolutional neural networks (CNNs) can automatically extract features from images depicting tea leaves affected by insect infestation and diseases (Xue et al., 2023). While traditional CNN uses convolution-based architecture, VIT employs transformer-based architecture, which has proven highly effective in natural language processing tasks (Alzahrani et al., 2023). Recently, vision transformers have seized the spotlight, captivating researchers with their remarkable prowess in addressing classification challenges across a spectrum of vision-based applications (Thakur et al., 2022). ViT follows the established data flow pattern of transformers, which eases its integration with different data types (Li et al., 2023). The transformer's acyclic network structure and parallel computing through encoder-decoder and self-attention mechanisms notably reduce training time and enhance machine translation performance (Zhan et al., 2023). Their extraordinary performance has spurred investigations into the potential of these transformative models within the domain of plant pathology (Thakur et al., 2023). This marks a notable progression in visual-based deep learning, where vision transformers have exhibited substantial promise in tasks extending beyond mere classification (Zhan et al., 2023).

The development of Vision Transformer (ViT) has spurred research in image recognition, facilitating computerized categorization and identification of tea leaf diseases (Gayathri et al., 2020). Vision transformer (ViT) is an essential technology for accurately detecting plant diseases and insect pests and ensuring system performance (Li et al., 2022). In precision tea production, it is crucial to identify diseases accurately and promptly. Different studies have developed automated systems using various imaging modalities to predict leaf diseases. One solution that has shown great potential for detecting tea leaf diseases is vision transformation. This method has proven highly accurate and efficient in identifying such diseases. As a result of its success, the vision transformer (ViT) has become the preferred approach for identifying plant diseases (Alzahrani et al., 2023; Mustofa et al., 2023; Yu et al., 2023). Vision transformer (ViT) has dramatically increased recently, owing to its remarkable performance in plant leaf disease detection (Boukabouya et al., 2022). Despite their demonstrated capabilities, adopting vision transformers in plant pathology applications remains an emerging and intriguing frontier (Li et al., 2023). Moreover, leveraging a vision transformer as an automated method for leaf disease detection in images can greatly facilitate farmers in promptly and cost-effectively implementing effective control measures (Hossain et al., 2023). However, the widespread application of Convolutional Neural Networks (CNNs) in plant disease detection has led to the consideration of a deep CNN with multiple hidden layers in our proposed research. This deep architecture aims to enhance disease classification in tea leaves by enabling the network to detect a larger number of features, ultimately resulting in improved accuracy in disease detection (Datta & Gupta, 2023). Unhealthy leaves display variations in color, texture, and size compared to healthy leaves, creating an opportunity to conduct image analysis through a CNN network. This analysis helps gather insights about pixel discrepancies across the entire leaf. Normally, the pixels in a leaf are expected to share common characteristics like color, intensity, and texture. Nonetheless, when a small cluster of pixels diverges from the norm, it indicates inconsistencies within the object or the existence of other elements (Ahad et al., 2023). This is achieved by integrating diverse attention levels and impacting the classification of plant leaf diseases within image data through various feature variations (Hossain et al., 2023).

Tea leaf diseases have a significant impact on the global tea industry. To address this, vision transformers in intelligent agriculture solutions are being developed for early disease detection and control (Thakur et al., 2022). Plant diseases and pests are a major problem for tea growers worldwide, resulting in significant tea loss. This affects the quality, nutritional value, and crop yield (Kalaydjian & C. T., 2023; Zhan et al., 2023). The biggest threat to the integrity of the tea supply is tea leaf diseases, which can significantly impact the output and quality of the crop. Therefore, it is crucial to identify and diagnose these diseases early to ensure successful treatment (Alzahrani et al., 2023). The presence of tea pests and diseases during tea harvesting are significant factors that can negatively impact tea production and quality. This can ultimately threaten macroeconomic stability and sustainable development (Fu et al., 2023). Moreover, detecting tea pests and diseases is challenging due to complex backgrounds and diverse leaf patterns. Traditional methods rely on expert knowledge, leading to inefficiencies and potential misclassifications (Wang et al., 2023). The occurrence of tea leaf diseases can influence the impregnability and quality of tea output. Identifying these diseases accurately can be challenging for workers, making detecting and diagnosing them crucial (Yu et al., 2023). Tea production has been negatively affected by diseases like tea algae leaf spot (TALS), tea bud blight (TBB), tea white scab (TWS), and tea leaf blight (TLB), resulting in a decrease in output (Bao et al., 2022). It is essential to increase tea production to prevent price hikes and alleviate the financial losses experienced by business owners (Thakur et al., 2021). Early detection of tea leaf diseases can help minimize the loss of tea yield (Mukhopadhyay et al., 2021). Diseases affecting tea leaves can lead to stunted growth in tea plants, resulting in a lower yield and inferior quality of tea leaves. These diseases cause a decrease of around 20% in the tea leaf yield annually, causing significant financial losses for tea farmers (Hu et al., 2019). Moreover, diseases and insect pests of tea leaves cause substantial economic losses to the tea industry every year, so the accurate identification of them is significant (Xue et al., 2023). Additionally, continuous pathogen exposure can lead to substantial crop yield losses in tea plants, making early disease detection essential to mitigate the impact on tea production. In a parallel vein, plant diseases can harm the agricultural sector by causing crop and economic losses. Therefore, timely diagnosis is crucial for effective disease management and control (Datta & Gupta, 2023; Tabbakh & Barpanda, 2023).

The aim is to identify plant diseases from images of leaves by developing a Vision Transformer-based deep learning technique (Thakur et al., 2021). A significant challenge to identifying tea leaf diseases in their earliest stages is to reduce the potential for associated economic damage (Alzahrani et al., 2023). Tea leaf diseases can reduce the quality of tea and cause severe economic losses to tea farmers. Accurate detection and identification of tea leaf diseases and timely prevention and control measures are significant to reduce the loss of tea production, improve tea quality, and increase tea farmers' income (Bao et al., 2022). Hence, this study is expected to minimize the workload of entomologists and aid in the rapid identification and detection of tea leaf diseases, thus minimizing economic losses (Soeb et al., 2023).

## Process of Literature Review:

Studies related to Vision Transformer for detection, feature extraction, feature extraction using traditional machine learning techniques for the classification, autoencoders (AEs), CNN (Convolutional Neural Network), and hybrid models for deep learning techniques included in this study. Most importantly, the research that experimented using a proper research methodology without providing experimental research was included in the literature review.

**Article Identification:**

The literature review of this study followed Systematic Reviews and Meta-Analyses (PRISMA) guidelines. A review of around 30 selected papers is presented in this paper. All the articles cover detecting tea leaf disease using Vision Transformer-based deep learning techniques.

In the literature review, the study identified preliminary sources using online databases. Primarily, Google Scholar was selected. The process of article selection is described below:

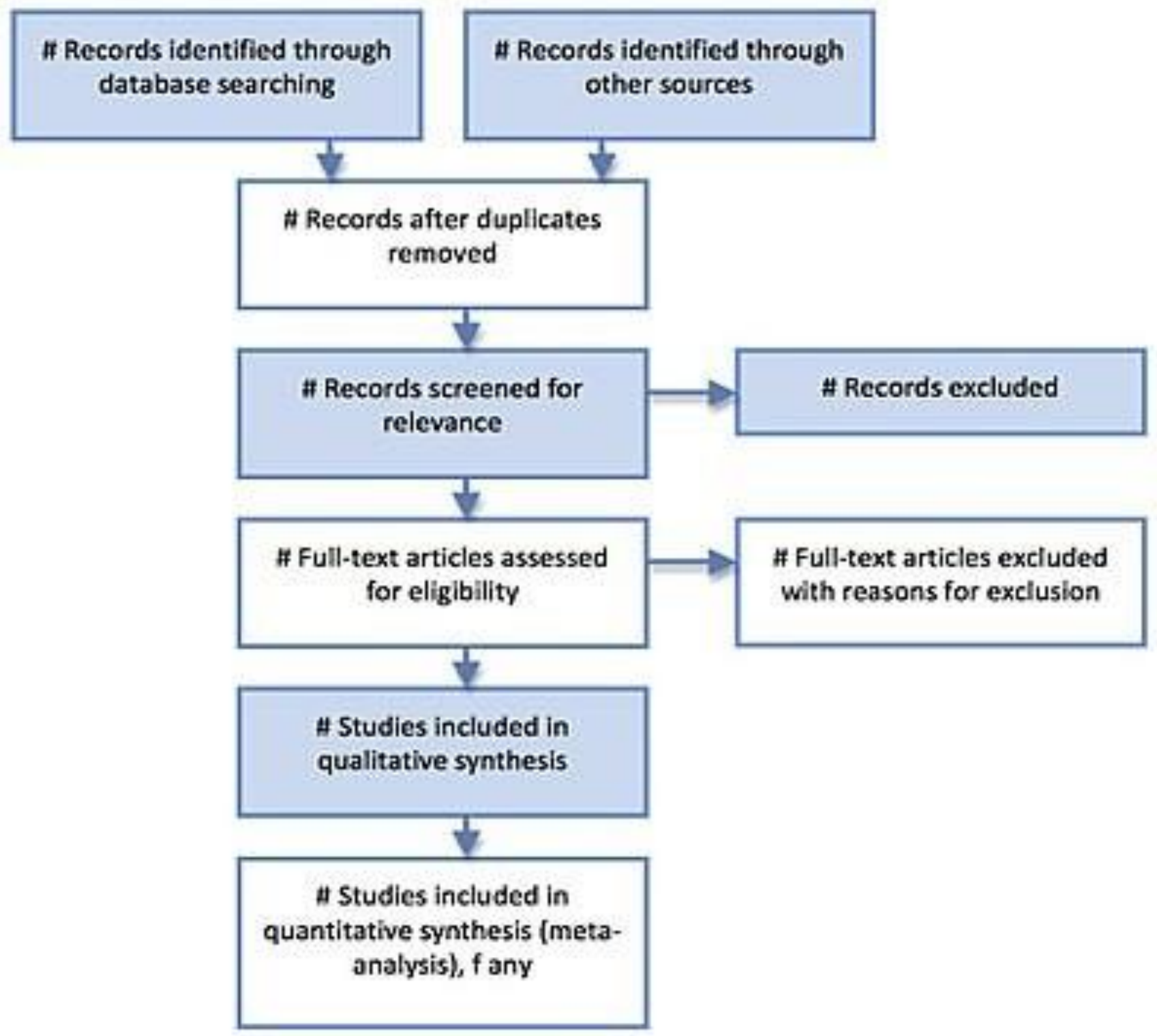

Figure 1: The Process of article selection

**Article Selection:**

Articles were selected for final review using a three-stage screening process based on inclusion and exclusion criteria. After removing duplicate records that were generated from using a database, articles were first screened based on the title alone. The abstract was assessed, and finally, the full articles were checked to confirm eligibility. The chief investigator conducted the entire screening process.

To meet the inclusion criteria, articles had to:

- Be original research articles published in a peer-reviewed journal with full-text access offered within our university.
- Involve the use of plant leaf images.
- Be published in English.
- Be concerned with applying machine learning techniques for plant leaf disease detection.
- Included articles were limited to those published from 2019 to 2023 to focus on deep learning methodologies. Here, a study was defined as work that employed a Vision Transformer-based deep learning algorithm to detect tea leaf disease and that involved the use of one or more of the following performance metrics: accuracy, the area under the receiver operating characteristics curve, sensitivity, specificity, or F1 score.

The exclusion criteria were:

- Review articles.
- Book or book chapters.
- Short communications or case reports.
- Unclear descriptions of data.
- No validation was performed.

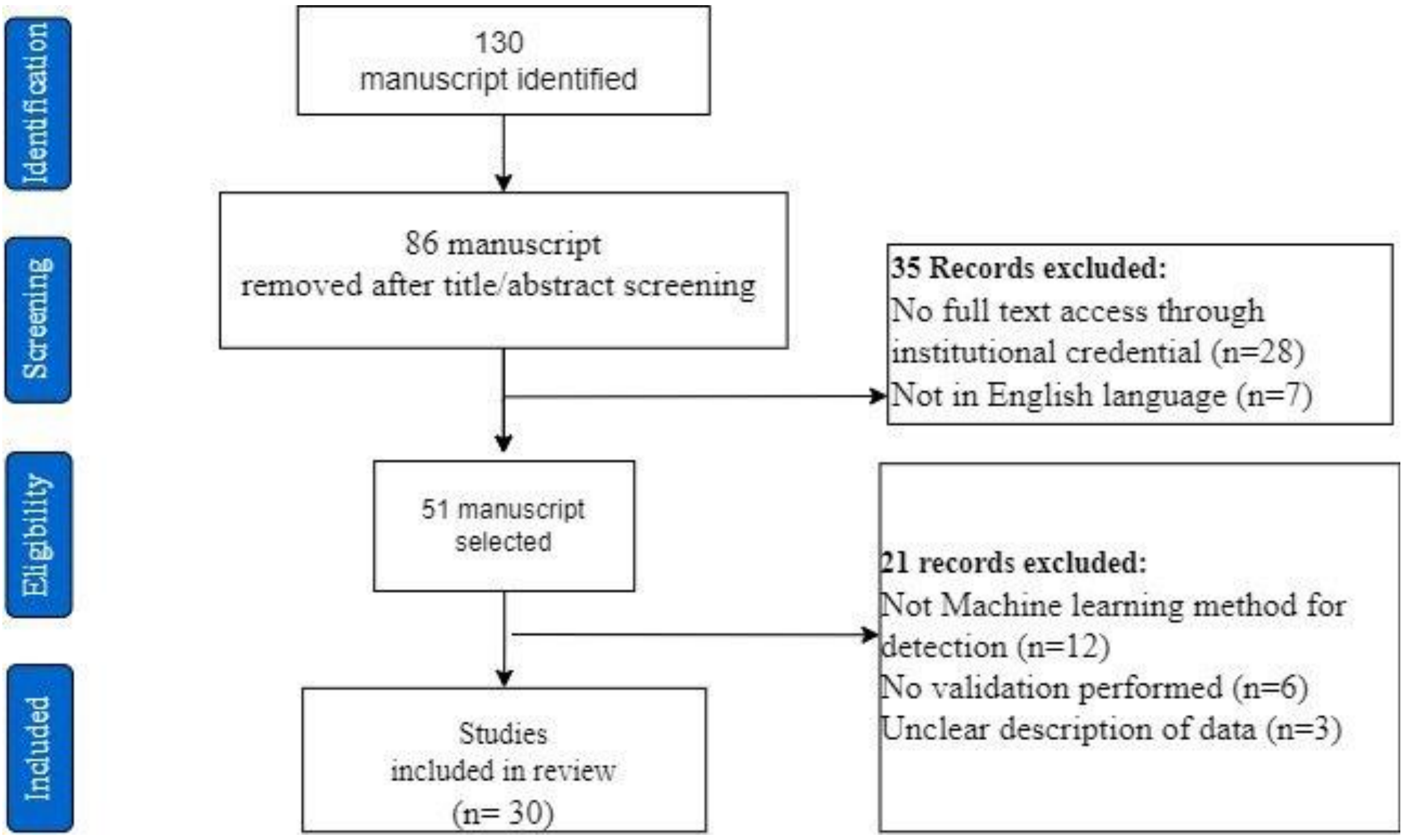

Figure 2: The Preprint Studies

## Literature Review:

The first group of researchers used the conventional Vision Transformer (ViT) to detect tea leaf diseases automatically at the field condition, which mitigates the economic loss. They employed Vision Transformer (ViT) because of its superior accuracy, adaptability to multispectral data, and scalability. ViT was excellent in capturing complex image patterns, making them invaluable for precise plant disease diagnosis. Due to its versatility and efficiency, authors preferred it over conventional CNNs in plant disease categorization. Among these researchers, Boukabouya et al. (2022) had the highest accuracy of 99.7%, whereas Silva & Brown (2023) had the lowest accuracy of 90.02%. However, Kalaydjian & C. T. (2023) also had a good accuracy of 99% in detecting plant leaf disease detection. Though Boukabouya et al. (2022) and Kalaydjian & C. T. (2023) got good accuracy, they used a publicly available dataset from PlantVillage. These studies were mainly conducted using only one dataset, and none of these authors focused on tea leaf datasets. However, it is unknown whether the model will perform better in other datasets, especially in tea leaf datasets.

The second group of researchers has consistently chosen advanced Vision Transformers (ViTs)

due to their superior performance and innovative features in plant disease detection. Here, Parez et al. (2023) proposed GreenViT, which shows ViTs' improved accuracy with convergence in a minimal number of epochs. Similarly, a hybrid model named TLMViT was proposed by Tabbakh and Barpanda (2023), which also achieved high validation accuracies. Another study by Zhu et al. (2023) introduced MSCVT by merging CNN and ViT, which integrates self-attention and multiscale convolution techniques for crop disease detection. Though the authors achieved good accuracy, they all used publicly available secondary datasets for their studies. Moreover, they did not do any experiments on tea leaf datasets. An Inception Convolutional Vision Transformer (ICVT) demonstrated by Yu et al. (2023) enhanced data richness by including inception architecture and cross-channel feature learning. It resulted in impressive accuracy rates on PlantVillage and ibean datasets but an average accuracy on AI2018 and PlantDoc datasets. Zhan et al. (2023) and Zhang et al. (2023) proposed IterationVIT and IEM-ViT for knowledge acquisition and meeting the specific requirements of diagnosing tea diseases. Both authors focused on tea leaf and used primary datasets in their work; unfortunately, they didn’t achieve higher accuracy in detecting tea leaf disease. Thakur et al. (2021) proposed PlantViT, which employed transformer-based techniques for automated, highly accurate plant disease identification. Furthermore, Thakur et al. (2022) introduced PlantXViT, which combines the best aspects of traditional CNNs and ViTs, proving effective across various datasets. Though they had achieved better accuracy in their work by using multiple datasets, they did not shed light on the tea leaf disease detection field.

Another group of researchers has advanced the field of plant leaf disease detection by developing ensembled models that combine vision transformers and other models. Their primary objective is to enhance the efficiency and accuracy of disease detection in agriculture, where early diagnosis is critical for crop yield and quality. Notable achievements include Li et al. (2023) with their SLViT hybrid network, reaching 98.84% accuracy on the Plant Village dataset, and Thakur et al. (2023) attaining 98.86% accuracy on 'PlantVillage.' However, Hossain et al. (2023) introduced a cutting-edge transformer-based model for tomato disease detection, with MaxViT achieving the highest accuracy at 97%. Alshammari et al. (2022) demonstrated 97% accuracy in binary and 96% in multiclass classification for olive leaf disease. Moreover, a comparison of three deep learning models, namely, DenseNet169, ResNet50V2, and ViT, is presented by Alzahrani et al. (2023). The study shows that DenseNet169 had a higher accuracy of 99% than the other two models. Though the authors achieved good accuracy in their work, the fact is that they did not use the primary dataset and did not center their attention on tea leaf disease.

Meanwhile, some researchers showed advancements in tea leaf disease detection by introducing advanced convolutional neural network (CNN) models. Datta and Gupta (2023) designed a deep CNN model to categorize diseased tea leaves into distinct groups, enhancing disease detection accuracy by recognizing a wide range of features, culminating in an impressive 96.56% accuracy. Their method identified specific tea leaf diseases, including Algal Spot at 98.23%, Brown Blight at 97.98%, and more. In contrast, Hu & Fang (2022) introduced MergeModel, which was conducted to detect tea leaf diseases in limited samples by combining multiple CNN modules, achieving a good performance in small sample sizes. However, Hu et al. (2019) improved the deep CNN method designed for tea leaf disease identification. Their changes included a feature extraction part that can look at different picture details, resulting in an average identification accuracy of 92.5%. While these contributions have significantly advanced the field, they also make us think about how well the method works in different situations and if it's practical to use in real-life farming settings where resources might be limited.

Moreover, some researchers shed light on tea leaf disease detection by developing YOLO-based models that have revealed object detection techniques' effectiveness. Among them, Soeb et al. (2023) showcased the high efficacy of the YOLOv7 approach, achieving an impressive value for detection accuracy (97.3%), precision (96.7%), recall (96.4%), mAP (98.2%), and F1-score (0.965) in identifying tea leaf diseases. Xue et al. (2023) introduced a model named YOLO-Tea that extends the capabilities of YOLOv5 by incorporating self-attention, convolution, receptive field blocks, and global context networks, leading to significant enhancements in the detection of tea tree leaf diseases and insect pests. In parallel, Bao et al. (2023) introduced DDMA-YOLO, which is a UAV-based remote sensing technique with RCAN and Dual-Dimensional Mixed Attention (DDMA) to increase the detection efficiency of Tea Leaf Blight (TLB), achieving significant improvements in AP0.5 and recall for TLB monitoring. These studies collectively demonstrated the promise of YOLO-based object detection techniques in tea leaf disease detection. However, it is essential to recognize that their practical utility may encounter obstruction when applied to real-world datasets.

Some other researchers lead the way in developing innovative methods for detecting plant leaf diseases using advanced technologies to tackle the specific challenges within the agricultural sector. Faisal et al. (2023) introduced a hybrid approach that combined MobileNetV3 and Swin Transformer models for coffee leaf disease detection, achieving a notable 84.29% accuracy. Meanwhile, Moyazzoma et al. (2021) performed a commendable 90.38% validation accuracy in classifying significant crop diseases in Bangladesh by employing convolutional neural networks and MobileNetv2 for feature extraction. Wang et al. (2023) demonstrated the potential of the YOLOv5 model enhanced with GAM and CBAM attention mechanisms, resulting in an average accuracy of 79.3% when detecting pests and diseases in tea plants, and aided further by the WBF algorithm. Thai et al. (2023) improved their transformer-based leaf disease detection model using optimization strategies, hinting at the prowess of advanced models for agricultural applications. Salamai et al. (2023) developed a lesion-aware visual transformer for detecting paddy leaf diseases, achieving an outstanding average accuracy of 98.74% and an f1-score of 98.18%. Lastly, Mukhopadhyay et al. (2021) proposed a Non-dominated Sorting Genetic Algorithm (NSGA- II) to detect diseased areas in tea leaves through image clustering, with an average accuracy of 83% for five distinct tea leaf diseases. However, none of these papers used the primary dataset for the experiment, so it's crucial to understand how to use these new ideas in real-life datasets and ensure they work well on a bigger scale.

## Table 1: Research Matrix

| Author | Model | Accuracy | Contribution |
|---|---|---|---|
| Kalaydjian & C. T. (2023) | ViT | 99% | Proposed pre-trained Vision Transformer (ViT) for plant disease classification, aiming to mitigate crop losses and ensure food security. |
| Silva & Brown (2023) | ViT | Training: 93.71 %<br>Test: 90.02% | Showed the practical utility of multispectral images and Vision Transformer (ViT) models for plant disease identification in field conditions. |
| Li et al. (2022) | ViT | 96.71% | Trained the constructed ViT network for automatically classifying plant diseases and insect pests. |
| Boukabouya et al. (2022) | ViT | 96.7%, 98.52%, 99.1% & 99.7%. | Proposed Vision Transformers (ViT) to detect tomato diseases at the leaf stage with high accuracy (up to 99.7%), enabling early interventions for plant health preservation and sustaining the natural cycle. |
| Thai et al. (2021) | ViT | 1% higher than CNN | Exploited Vision Transformer (Vit) in place of a convolution neural network (CNN) for detecting cassava leaf diseases. |
| Parez et al. (2023) | GreenViT | 100% on PV, 98% on DRLI & 99% on PC dataset | Introduced a fine-tuned GreenViT model using Vision Transformers (ViTs) for plant disease detection. |
| Tabbakh & Barpanda (2023) | Transfer Learning Model & Vision Transformer (TLMViT) | 98.81% on PlantVillage & 99.86% wheat dataset | Proposed TLMViT, a hybrid model for plant disease classification, supported by extensive experiments and demonstrating the exceptional potential of |
| | | | vision transformers for profound feature extraction. |
| Zhu et al. (2023) | MSCVT | 99.86% on PlantVillage & 97.50% on Apple Leaf Pathology | Introduced MSCVT, a hybrid model merging CNN and vision transformer, which integrated multiscale self-attention and inverted residual blocks for crop disease recognition. |
| Yu et al. (2023) | Inception Convolutional Vision Transformer (ICVT) | 99.94% on VillagePlant, 99.22% on ibean, 86.89% on AI2018 & 77.54% on PlantDoc. | Proposed automatic plant disease identification based on inception convolution and vision transformer. |
| Zhan et al. (2023) | IterationVIT | Classification: 98% F1-score: 96.5%. | Introduced IterationVIT model for diagnosing tea diseases, which blends local feature extraction using convolution and global feature extraction via transformers. |
| Zhang et al. (2023) | Information Entropy Masked Vision Transformation (IEM-ViT) | 93.78% | Improved the recognition accuracy by nearly 20% compared to the currently standard image recognition algorithms, including the ResNet18, VGG16, and VGG19. |
| Thakur et al. (2022) | PlantXViT | 93.55% on Apple, 92.59% on Maize & 98.33% on Rice | Proposed a model that combines the capabilities of traditional CNN with Vision Transformers to efficiently identify plant diseases for several crops. |
| Thakur et al. (2021) | PlantViT | 98.61% on PlantVillage & 87.87% on Embrapa | Introduced "PlantViT," a hybrid CNN-Vision Transformer model for plant disease detection, aiming to enhance agricultural production and address food security challenges. |
| Li et al. (2023) | SLViT (SHDC + LViT block) | 98.84% on PlantVillage & 87.64% on SLD10k dataset | Proposed a hybrid network called SLViT combining flexible transformers (LViT) and lightweight CNN architecture (SHDC) for highly efficient and precise disease recognition. |
| Thakur et al. (2023) | ViT + CNN | 98.86% on PlantVillage & 89.24% on Embrapa | Introduced a hybrid model by combining vision transformers and CNNs for effective disease identification of plant leaves and offered a lightweight structure suitable for IoT-based agricultural systems. |
| Hossain et al. (2023) | External Attention Transformer (EANet), Multi-Axis Vision Transformer (MaxViT), Compact Convolutional Transformers (CCT) & Pyramid Vision Transformer (PVT) | EANet: 89%<br>MaxViT: 97%<br>CCT: 91%<br>PVT: 93% | Presented four cutting-edge transformer-based models (EANet, MaxViT, CCT, and PVT) and highlighted that MaxViT is the most effective model for tomato leaf disease classification. |
| Alshammari et al. (2022) | CNN + ViT | 96% for multiclass classification & 97% for binary classification. | Developed a unique deep ensemble learning strategy that combines CNN with a vision transformer (ViT) model. |
| Alzahrani et al. (2023) | DenseNet169, ResNet50V2 & ViT | 99.00%, 95.60% & 98.00% | Introduced a computer vision-based deep learning approach, comparing different models, such as DenseNet169, ResNet50V2, and ViT, to accurately classify tomato leaf diseases for early detection. |
| Datta & Gupta (2023) | Deep CNN | 96.56% | Presented a Deep CNN with multiple concealed layers to categorize tea leaf diseases, improving accuracy. |
| Hu & Fang (2022) | MergeModel [multi-convolutional neural | | Proposed MergeModel, a multi-convolutional |
| | network (CNN) model] | | neural network for tea leaf disease identification, and SinGAN-based data augmentation to achieve higher accuracy in small sample settings. |
| Hu et al. (2019) | Deep CNN | 92.5% | Improved a deep convolutional neural network to identify tea leaf diseases |
| Soeb et al. (2023) | YOLOv7 | Detection: 97.30%<br>Precision: 96.70%<br>Recall: 96.40%<br>mAP: 98.2%<br>F1-score: 0.965 | Proposed tea leaf disease detection and identification based on YOLOv7. |
| Xue et al. (2023) | YOLOv5 + CBAM + ACmix + RFB + GCnet<br>(YOLO-Tea) | $AP_{0.5}$: 79.3%<br>$AP_{TLB}$: 73.7%<br>$AP_{GMB}$: 82.6% | Presented YOLO-Tea, an advanced YOLOv5-based model enriched with self-attention, convolution, RFB module, and GCNet to address tea leaf diseases better. |
| Bao et al. (2023) | Unmanned Aerial Vehicle (UAV) + DDMA-YOLO | Enhancement:<br>$AP_{0.5}$: 3.8%,<br>Recall: 6.5% | Combined UAV remote sensing with DDMA-YOLO for TLB (Tea Leaf Blight) detection and monitoring. |
| Faisal et al. (2023) | MobileNetV3 + Swin Transformer | 84.29%. | Introduced a hybrid feature fusion approach for coffee leaf disease identification, combining |
| | | | MobileNetV3 and Swin Transformer models. |
| Moyazzoma et al. (2021) | MobileNetv2 | 90.38% | Proposed a MobileNetv2-based model to classify leaf diseases in key crops of Bangladesh, addressing economic reliance on agriculture. |
| Wang et al. (2023) | CBAM_fusion_GAM | 79.3% | Enhanced tea pest and disease detection by introducing GAM and CBAM attention mechanisms into YOLOv5. |
| Thai et al. (2023) | FormerLeaf | 3% enhancement | Presented FormerLeaf, a transformer-based detection model enriched with LeIAP and SPMM optimizations. |
| Salamai et al. (2023) | Lesion-Aware Visual Transformer | Accuracy: 98.74%<br>f1-score: 98.18% | Presented a novel approach for accurately detecting paddy leaf diseases using a lesion-aware visual transformer and innovative feature extraction methods. |
| Mukhopadhyay et al. (2021) | Non-dominated Sorting Genetic Algorithm (NSGA-II) | 83% | Proposed NSGA-II based image clustering for detecting the disease area in tea leaves. |

## Inference from current research studies:

According to the research matrix (Table 1), there has been a recent upswing in plant disease detection research, fueled by the integration of cutting-edge machine learning and transformer-based models. Within the domain of tea leaf disease detection, a variety of research has been undertaken by different scholars, employing a wide range of methodologies and models. The initial group of researchers, exemplified by Boukabouya et al. (2022) and Kalaydjian & C. T. (2023), have showcased the potential of traditional Vision Transformers (ViTs) owing to their exceptional precision and capacity to adapt to multispectral data. While Boukabouya et al. attained a remarkable accuracy of 99.7%, the overarching concern pertains to their limited emphasis on tea leaf datasets, prompting questions about the broader applicability of ViTs in this specific domain. Other authors like Parez et al. (2023), Tabbakh and Barpanda (2023), and others consistently opted for advanced ViTs in the context of plant disease detection, achieving commendable validation accuracy. However, the absence of experiments conducted on tea leaf datasets within this category introduces uncertainties about the viability of these models for tea leaf disease detection.

Vision Transformers are combined with other models to boost the efficiency and accuracy of disease detection. Noteworthy authors in this category, such as Li et al. (2023) and Thakur et al. (2023), have presented encouraging results in the context of standard plant disease datasets. However, the absence of a primary dataset and a specific focus on tea leaf diseases in these studies indicates the necessity for further investigation. Simultaneously, researchers have significantly contributed to the field by developing deep Convolutional Neural Network (CNN) models explicitly tailored for tea leaf diseases. As demonstrated by authors like Datta and Gupta (2023), these models have shown impressive accuracy in identifying various tea leaf diseases. Nevertheless, their practicality in a wide range of real-world farming scenarios remains an area that requires further examination.

Furthermore, the review of existing literature emphasizes the potential of YOLO-based models for object detection in identifying tea leaf diseases, as evidenced by research conducted by Soeb et al. (2023) and other studies, which have demonstrated exceptional accuracy and precision in detecting these diseases. However, the applicability of these models to real-world datasets requires further careful examination. Additionally, a group of researchers has introduced innovative techniques and technologies, including MobileNet, Swin Transformer, and advanced YOLO models, showcasing commendable accuracy in various agricultural applications. Nevertheless, the absence of experiments using the primary dataset underscores the need for more comprehensive investigations into their practicality and effectiveness in tea leaf disease detection. In summary, these insights underscore the dynamic nature of research in tea leaf disease detection and emphasize the essential role of future research in addressing the unique challenges and datasets pertinent to this crucial field.

## Limitations:

The objective of this study is to conduct a comprehensive assessment of the current landscape of tea leaf disease detection through the application of deep learning techniques. Additionally, the study aims to shed light on a relatively less explored area in this field, which involves the utilization of Vision Transformers for addressing the losses in yield and quality of tea. This research direction is closely aligned with the overarching objective of harnessing technological advancements to improve disease management strategies and enhance tea production.

It is noteworthy that a significant portion of the existing literature revolves around the utilization of publicly available and freely accessible datasets for training and evaluating models designed for detecting diseases in tea leaves. Most of the papers are mainly conducted using one dataset. However, it is not known how the model will perform in other datasets. Moreover, it is crucial to acknowledge that the effectiveness of deep learning methods in disease classification heavily relies on the availability of substantial volumes of well-annotated images. Acquiring a diverse and extensive dataset containing images accurately labeled with instances of various diseases poses a considerable challenge.

Creating datasets like these is quite a complex and time-consuming process. Tea leaf diseases come in many forms, and factors such as lighting and growth stages can change how they appear, which makes building a reliable dataset challenging. Training deep learning models effectively requires diverse and high-quality training data, which becomes difficult due to these challenges. People who specialize in studying tea leaf diseases face gathering many pictures with accurate labels to identify different diseases in each image. This demands expertise and a significant investment of time. As a result, many researchers choose to work with existing datasets accessible to everyone.

## Direction for Future Research

The future research directions for improving the accuracy and efficiency of tea leaf disease detection using Machine learning include reinforcement learning, hybrid machine learning, and case-based reasoning. Reinforcement learning can be used to develop a ViT model that can automatically identify diseased tea leaves without the need for human intervention. Hybrid machine learning can combine several models with a traditional machine learning algorithm to improve the accuracy of tea leaf disease detection. Case-based reasoning can be used to develop a machine-learning model that can learn from the mistakes of previous models and improve its performance over time.

In addition to these novel research directions, several other challenges must be addressed to improve the accuracy and efficiency of tea leaf disease detection using ViT. These challenges include data scarcity, environmental variability, and real-time detection. Despite these challenges, ViT is a promising technology for tea leaf disease detection. By continuing to research novel techniques and address the existing challenges, it is possible to develop ViT models that can accurately and efficiently detect tea leaf diseases.

## Conclusion:

This study serves as a comprehensive review of tea leaf disease detection, exploring various approaches and methodologies for early diagnosis and management of plant diseases. Vision Transformers (ViT) have emerged as a promising technology for accurately classifying and identifying tea leaf diseases alongside other models like deep convolutional neural networks (CNN), YOLO-based techniques, and hybrid models. Researchers have assessed the strengths and weaknesses of these models and conducted in-depth evaluations using metrics like accuracy, precision, and recall. While ViT-enabled CNN models and other innovative techniques show potential, it's imperative to consider their real-world applicability, particularly in the context of tea leaf datasets. Additionally, ensembled models combining vision transformers and other approaches have demonstrated their efficiency, while advanced CNN models and YOLO-based models have proved effective in recognizing various tea leaf diseases. These findings collectively offer valuable insights into the advancement of tea leaf disease detection, with a focus on enhancing agricultural practices and addressing the specific challenges faced by farmers in real-life settings.